\documentclass[conference]{IEEEtran}
\IEEEoverridecommandlockouts
\usepackage{cite}
\usepackage{amsmath,amssymb,amsfonts}
\usepackage{algorithmic}
\usepackage{graphicx}
\usepackage{textcomp}
\usepackage{tabularx, booktabs}

\usepackage{tikz}
\usepackage{pgfplots}
\usepackage{import}
\usepackage{mathbbol}
\usepackage{bbold}
\usepackage{float}
\usepackage{environ}
\usepackage{pgfplotstable}
\usepackage{setspace}
\usepackage{hyperref}
\usepackage{soul}
\usepackage[bottom]{footmisc}
\usepackage{lipsum}

\pgfplotsset{compat=1.18}
\usepackage{balance}
\usepackage[colorinlistoftodos]{todonotes}

\usepackage{balance}
\usepackage{xcolor,colortbl}
\usepackage[colorinlistoftodos]{todonotes}
\def\BibTeX{{\rm B\kern-.05em{\sc i\kern-.025em b}\kern-.08em
    T\kern-.1667em\lower.7ex\hbox{E}\kern-.125emX}}

\begin{document}

\title{Transfer Entropy in Graph Convolutional Neural Networks\\
}

\author{
    \IEEEauthorblockN{Adrian Moldovan\IEEEauthorrefmark{1}\IEEEauthorrefmark{2},
        Angel Ca\c taron\IEEEauthorrefmark{1}\IEEEauthorrefmark{2},
        R\u azvan Andonie\IEEEauthorrefmark{3}\IEEEauthorrefmark{1}}
    
    \IEEEauthorblockA{\IEEEauthorrefmark{1}\textit{Department of Electronics and Computers}, Transilvania University, Bra\c{s}ov, Romania
    }
    
    \IEEEauthorblockA{\IEEEauthorrefmark{2}
        Siemens Technology, Siemens SRL, Bra\c sov, Romania
    }
    
    \IEEEauthorblockA{\IEEEauthorrefmark{3}Central Washington University, Ellensburg, WA, USA\\
        Email: adrian.moldovan@gmail.com, cataron@unitbv.ro, razvan.andonie@cwu.edu
    }
}

\maketitle

\begin{abstract}
Graph Convolutional Networks (GCN) are Graph Neural Networks where the convolutions are applied over a graph. In contrast to Convolutional Neural Networks, GCN's are designed to perform inference on graphs, where the number of nodes can vary, and the nodes are unordered. In this study, we address two important challenges related to GCNs: \emph{i)} oversmoothing; and \emph{ii)} the utilization of node relational properties (i.e., heterophily and homophily). Oversmoothing is the degradation of the discriminative capacity of nodes as a result of repeated aggregations. Heterophily is the tendency for nodes of different classes to connect, whereas homophily is the tendency of similar nodes to connect. We propose a new strategy for addressing these challenges in GCNs based on Transfer Entropy (TE), which measures of the amount of directed transfer of information between two time varying nodes. Our findings indicate that using node heterophily and degree information as a node selection mechanism, along with feature-based TE calculations, enhances accuracy across various GCN models. Our model can be easily modified to improve classification accuracy of a GCN model. As a trade off, this performance boost comes with a significant computational overhead when the TE is computed for many graph nodes. 
\end{abstract}


\section{Introduction}

Graph Neural Networks (GNNs) are powerful deep learning models tailored to process data represented in graph form, encapsulating features as nodes and their interactions through edges for a wide array of domains, from social network analysis to protein compound interactions. GNNs have evolved from the early attempts that aimed at extending neural network methodologies to graph-structured data, marked by significant milestones that include the introduction of recurrent GNNs \cite{scarselli2009}, spectral methods \cite{bruna2014} and the message-passing framework \cite{gilmer2017}. Using the new convolutional operator, the latter evolved into more sophisticated architectures, such as Graph Convolutional Networks (GCNs) \cite{kipf2017}, Graph Attention Networks \cite{velickovic2018}, and Graph Isomorphism Networks \cite{xu2019}, each introducing novel mechanisms for aggregating and transforming node features based on graph topology or node features and edge properties. Their success always relied on reduced training time as well as important accuracy gains compared with the existing methods when applied on the same graph based data.

Despite their success, GCNs faced inherent challenges arising from both data structure and its architectural training mechanisms. The most important challenge is oversmoothing, the result of repeated aggregation. This may degrade the discriminative capability especially of the high degree nodes. Such nodes exert powerful influence on their neighbors through message propagation, leading to a scenario where, in advanced stages of training, the nodes become susceptible to misclassification due to the diminished classification strength, stemming from excessive information blending. Addressing this challenge while looking to capture both node features and graph topology remains an active area of research, with ongoing advancements seeking to unlock the full potential of neural networks on graphs \cite{wu2021}.

A second challenge is related to the utilization of node relational properties: heterophily and homophily. Heterophily is the tendency for nodes of different classes to connect, whereas homophily is the tendency of edges to connect similar nodes. The GCNs model accuracy is heavily influenced by the presence of heterophily and homophily in the graph structure. One approach to mitigate the misclassification of certain nodes in the absence of actual labels involves utilizing data inferred from the nodes' neighbors. This method is commonly employed, but it encounters several problems. The propensity of a node for homophily or heterophily can evolve during training, with the attributes exhibiting continuous variation based on the node's classification cluster.

We attempt to answer these challenges with a new strategy, embedding in the GCN learning stage a control mechanism based on Transfer Entropy (TE), introduced as a measure of the amount of directed transfer of information between two random processes \cite{Schreiber2000}, and later applied to graphs representing neural networks \cite{Lizier2011, Herzog2017, Herzog2020, Obst2010, Moldovan2020, Moldovan2021, Moldovan2023, Duan2023}.

Our main contribution is the following. We use the TE value to directly modify the features of GCN nodes. This way, we control the exchanged messages between these nodes and, consequently, the misclassification rate. TE is computed prior to the convolution operations, and applied after the convolution blocks. Our findings indicate that updating a percentage of the nodes exhibiting the highest heterophilic characteristics can increase the accuracy of a GCN classifier. Furthermore, it proves to be unnecessary to compute TE at every iteration - periodic assessment across a subset of epochs is sufficient.

Section \ref{background} of the paper reviews the main notations and concepts used in GCNs and TE and also explains how we use TE in the context of GCNs. Section \ref{contribution} introduces our proposed method: the GCN with TE-controlled learning mechanism. In Section \ref{experiments} we describe and discuss a series of experiments. Section \ref{conclusions} concludes with the final remarks.

\section{Background: Graph Convolutional Neural Networks and Transfer Entropy} \label{background}

This section describes the message aggregation mechanism of the convolutional GCN operator, a key component of our approach. We also introduce the TE notations used in our approach, and refer to previous work related to the application of TE in neural networks. 

\subsection{Graph Convolutional Networks}

GNNs generalize the convolution operation from tabular data to graph data, enabling the model to learn by aggregating and transforming feature information from a node's neighborhood. The message passing (or message-based propagation) mechanism is central to this process, allowing information to flow across the edges of the graph, thereby capturing the structural dependencies within the data.

GCNs adopted the message passing mechanism from chemistry \cite{gilmer2017}. This was established as a first-class tool by \cite{hamilton2017, kipf2017}. Later, some researchers attempted to use the graph structure \cite{zhu2020, pei2020, wei2022, topping2022, cavallo2023, gao2023, chen2023, huang2024} instead of using the available features and weights, while others used a different message propagation mechanisms \cite{yan2022, luan2023}. Well-known instruments from deep neural networks, such as skip-connections, have been applied in \cite{chen2020}. Attention methods where used to identify and remove nodes with similar features to improve performance \cite{velickovic2018}. In \cite{platonov2023} the proposed technique removed edges in a manner analogous to how pruning mechanisms operate within neural networks.

In \cite{Duan2023} authors leveraged TE to deduce prior insights from input time series data, subsequently applying these insights to enhance forecasts through a GNN framework.

GCNs reached soon certain accuracy thresholds that were not exceeded due to well-known, in-design limitations, for which only partial solutions have been found. These limitations are revealed by the intrinsic properties of the datasets and by the natural organization of graph data. Among these properties that influence the GCNs performance are hetherophily and homophily \cite{zhu2023}.

The main problem generated by the convolution operator is the effect of oversmoothing of a node's features \cite{rusch2023}, in particular for high-degree nodes. Any attempts to solve this challenge proved to be only incremental improvements, efficient only for datasets sharing similar particularities. 

The studies \cite{yan2022, magner2020}, \cite{magner2021} and \cite{bodnar2022} have provided theoretical limits that bound the accuracy levels of GCNs within certain constraints, yet they fall short of offering concrete solutions to these limitations. Additionally, while certain datasets intrinsically exhibit homophilic or heterophilic characteristics, the predominant scenario involves a mixture of both node types, distributed across various cluster-like configurations. This diversity renders traditional message-passing and convolution-based approaches less effective. A further challenge identified in recent advancements pertains to the observation that, during advanced stages of training, a subset of nodes may exhibit a decline in classification accuracy, thereby negatively impacting the overall performance.

Understanding the complexities of GCN architectures demands sophisticated visualization tools beyond what is typically used for tabular data. Effective visualization \cite{ Kovalerchuk2022integrating, Kovalerchuk2024} include pre-rendering processes like clustering, community detection for node alignment, scalability to handle vast datasets, aligning and separating nodes and edges strategically for optimal comprehension and customizable layouts to ensure clarity. These pre-rendering steps are vital for revealing the graph's embedded patterns. We will apply such visualizations in Section \ref{experiments}.

\begin{figure}[H]
\hspace*{-2.8cm}
\includegraphics[width=390pt, keepaspectratio]{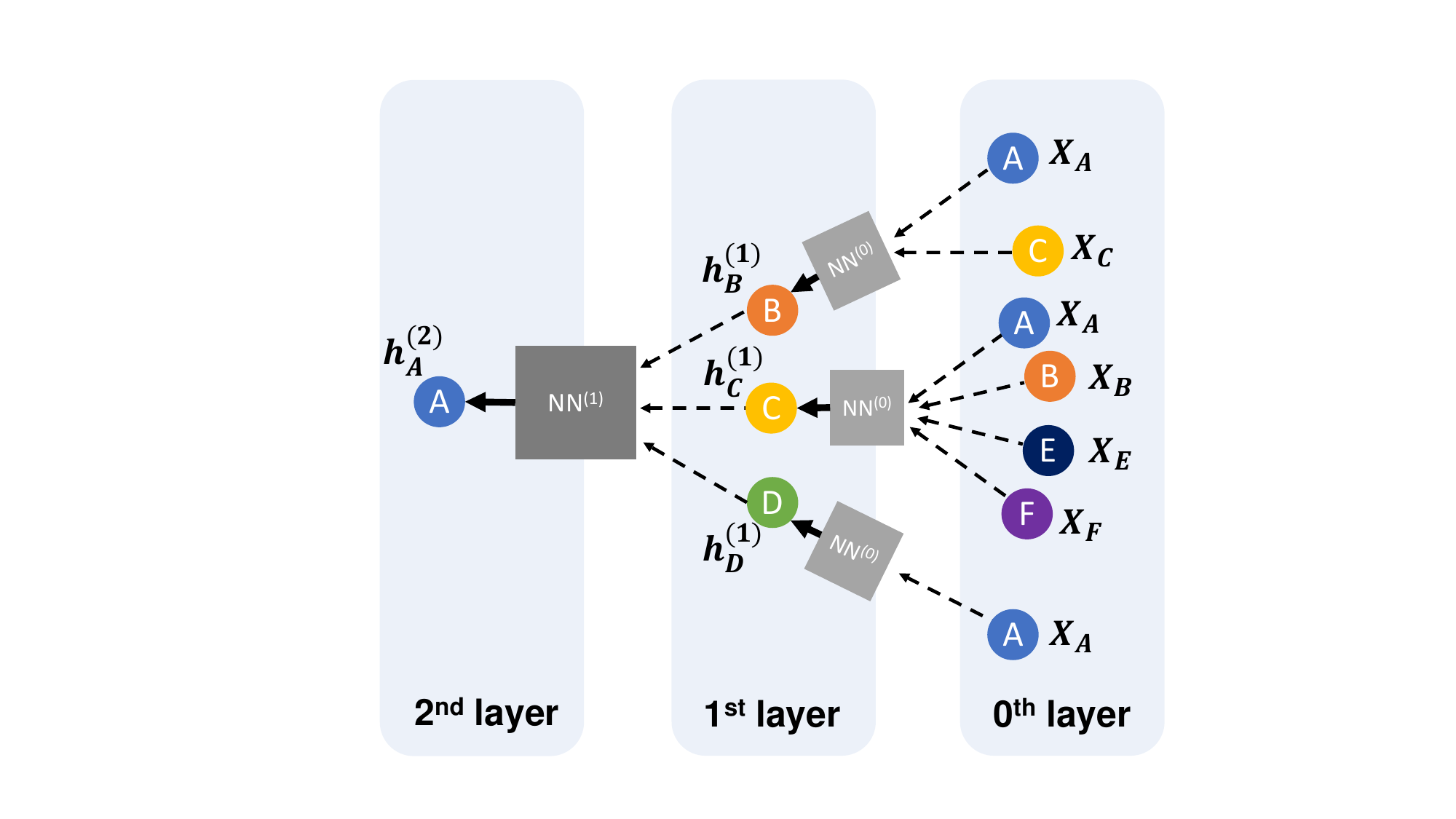}
\caption{Illustration of a node's features being aggregated through a two layer convolution in a GCN (graphic from \cite{yoon2022}). The node A from 2nd layer will receive updated aggregated values from all its neighbors having a depth equal with the number of convolutional layers.}
\label{image:messagepassing}
\end{figure}

At this step, we introduce some formalism. A graph can be defined as \(G = (V, E)\), where \(V\) is the set of nodes with \(|V| = N\) denoting the number of nodes and \(E \subseteq V \times V\) is the set of edges connecting the nodes. Each node \(i\) can have an associated feature vector \(x_i\), leading to a feature matrix \(X \in \mathbb{R}^{N \times F}\) for the graph, where \(F\) is the feature's vector dimension. In our datasets, \(F\) will be the length of the token vector that describes a single document from the citation dataset.

A fundamental operation in GNNs is the aggregation of neighboring node features, mathematically formalized as \[H^{(l+1)} = \sigma\left(AH^{(l)}W^{(l)}\right),\] where \(H^{(l)}\) denotes the node features at layer \(l\), $A \in \mathbb{R}^{N \times N}$ is the adjacency matrix of the graph, \(W^{(l)}\) is a learnable weight matrix, and \(\sigma\) represents a non-linear activation function.

Using the definition from \cite{kipf2017}, a convolutional layer within a graph aggregates the information of the neighbors for each node:

\begin{equation}
H^{(l+1)} = \sigma\left(\hat{D}^{-1/2} \hat{A} \hat{D}^{-1/2} H^{(l)} W^{(l)}\right),
\label{eq:convupdate}
\end{equation}
where \(H^{(0)} = X\), \(\hat{A} = A + I_N\) is the adjacency matrix with added self-loops (via the identity matrix \(I_N\)), \(\hat{D}\) is the diagonal degree matrix of \(\hat{A}\), and \(\hat{D}_{ii} = \sum_j \hat{A}_{ij}\). The process is illustrated in Fig. \ref{image:messagepassing}.

\subsection{Transfer Entropy}

TE serves as a pivotal statistical metric for assessing the synchrony or coherence between sequential events (typically, time series). This metric was associated with Granger's causality by several scholars \cite{Barnett2009, Hlavackova-Schindler2011}, albeit with a critical distinction. The application of "causality" in isolation is considered a misnomer. To mitigate ambiguity, Granger adopted the descriptor "temporally related" in 1977 \cite{granger1977}. While TE may reveal temporal correlations between variables, it does not serve as an unequivocal indicator of causality. Interpretations derived from TE analyses necessitate caution, particularly within the ambit of causality. 



For two random variables $X$ and $Y$, we can compute TE adapting Schreiber's \cite{Schreiber2000} formula:

\begin{equation}\label{eq:TEcond}
	TE_{Y\rightarrow X}=\sum_{t=1}^{n-1}{p(x_{t+1},x_{t}^{(k)},y_{t}^{(l)}) \: log \frac{p(x_{t+1}|x_{t}^{(k)},y_{t}^{(l)})}{p(x_{t+1}|x_{t}^{(k)})}} 
\end{equation}

In our prior research \cite{Moldovan2020}, we used TE as a modifier for the gradient within the backpropagation process of feedforward neural networks. This integration notably diminished the duration of the training phase and increased the stability of the model. Subsequently, our efforts were broadened to encompass Convolutional Neural Networks (CNNs). We employed \cite{Moldovan2021} TE as a regularization mechanism that enhances stability by being activated intermittently, instead of following the processing of each individual input, thereby ensuring a more controlled and efficient training regimen. We demonstrated \cite{Moldovan2023} that TE performs as a regulatory agent throughout the training stage of CNNs, offering additional compression techniques akin to those identified within the realm of information bottleneck theories, while also enhancing the performance of CNNs.

\section{The TE control mechanism in GCNs} \label{contribution}

This section describes our contribution - which consists of applying TE after the convolution block steps of GCNs. We name this novel architecture Transfer Entropy Generalized Graph Convolutional Network - TE-GGCN. Basically, it is an embedding of the TE control mechanism in the GGCN algorithm from \cite{yan2022}.

\subsection{The GGCN method}
We first revisit notations and core ideas of the GGCN model \cite{yan2022}, in which the dynamics of the nodes that are susceptible to misclassification are examined with respect to their interactions with adjacent nodes. The GGCN, method (including the implementation) synthesizes new ideas with previous GCN results. To decrease the rate of misclassification, two distinct strategies are proposed. In the first strategy, the weights of edges are recalibrated by enhancing the calculated relative degree of a node, expressed as $\bar{r_i}$, given that a node's actual degree, $r_i$, remains immutable:

\begin{equation}
\label{eq:ri}
\overline{r_i} = \mathbf{E}_{\mathbf{A} \mid d_i}\left(\left.\frac{1}{d_i} \sum_{j \in \mathcal{N}_i} r_{i j} \right\rvert\, d_i\right),
\end{equation}
where $r_{i j} = \sqrt{\frac{d_i+1}{d_j+1}}$ and $d_i$, $d_j$ are the degrees of nodes $i$ and $j$.

The second strategy involves the adjustment of edge features by assigning negative signs for heterophilous edges or positive signs for homophilous ones.

A convolutional layer is updated as follows:
\begin{equation}
\begin{split}
\mathbf{H}^{(l+1)}=\sigma(\hat{\alpha}^l(\hat{\beta}_0^l \widehat{\mathbf{H}^{(l)}}+\hat{\beta}_1^l(\mathbf{S}_{\text {pos}}^{(l)} \odot \widehat{\tilde{\mathbf{A}}^l}) \widehat{\mathbf{H}^{(l)}} + \\
\hat{\beta}_2^l(\mathbf{S}_{\text {neg}}^{(l)} \odot \widehat{\tilde{\mathbf{A}}^l}) \widehat{\mathbf{H}^{(l)}})), \\
\end{split}
\label{eq:eq1}
\end{equation}
where $\widehat{\mathbf{H}^l}=\mathbf{H}^{(l)}\mathbf{W}^{(l)} + \mathbf{b}^{(l)}$ is the $l$-th layer and $\sigma$ denotes the Elu activation function \cite{elu}. We apply the \emph{softmax} function \cite{McFadden1972Softmax} to the scalars $\hat{\beta}_0^l$, $\hat{\beta}_1^l$, and $\hat{\beta}_2^l$ obtained from the scalars ${\beta}_0^l$, ${\beta}_1^l$, and ${\beta}_2^l$. The latter are learned for each convolutional layer. The $\hat{\beta}^l$ scalars are behaving as scaling coefficients for the computed features in addition to the positive and negative coefficients. The sum of the $\hat{\beta}_i^l$ scalars is 1. Positivity of the \(\beta\) parameters must be enforced to not interfere with the $\mathbf{S}_{\text {neg}}^{(l)}$ edges sign matrix. The matrices $\mathbf{S}_{\text {pos}}^{(l)}$ and $\mathbf{S}_{\text {neg}}^{(l)}$ represent the positive and negative attention components, respectively, obtained from the sign matrix $\mathbf{S}^{(l)}[i, j] = \operatorname{cosine}(\mathbf{f}_i^{(l)}, \mathbf{f}_j^{(l)})$, which assigns signs to the edges between nodes based on the cosine similarity between their feature vectors at the $l$-th layer, denoted by $\mathbf{f}_i^{(l)}$ and $\mathbf{f}_j^{(l)}$.

The GGCN is augmented with a degree-based scaling mechanism, utilizing pre-computed node degree information coupled with a softplus transformation to dynamically adjust the influence of nodes based on their connectivity degrees. This preparatory step aims to enhance signal propagation and mitigate over-smoothing by tailoring node influence.

Subsequently, the model transforms input node features through a fully connected layer, enabling non-linear feature transformations. Later, node similarities are captured by applying pairwise product between input features and their transpose. These products are then normalized to limit numerical instabilities and eliminate self-similarities, fostering a robust representation.

The core of the GGCN method lies in its unique treatment of feature propagation via positive and negative attention matrices, derived from the adjacency matrix scaled by the degree-based coefficient. This bifurcated propagation obtained through signed attention mechanism is regulated by softmax-normalized coefficients and a softplus-scaled factor, combines positively and negatively influenced feature propagations with the original node features. This blending ensures a balanced consideration of both homophilic and heterophilic relationships within the graph.

\subsection{The proposed TE-GGCN method}

There is no general agreement on how homophily and heterophily should be computed. As suggested in \cite{platonov2024}, we use the following formula to compute the heterophily of a node $v$:
\begin{equation}
\begin{split}
  &\mathcal{H}_v=\frac{1}{|N(v)|} \sum_{u \in N(v)} 1\left(l_u \neq l_v\right)
\end{split}
  \label{eq:heterophily}
\end{equation}
where $|N(v)|$ denotes the number of neighbors node $v$ has, and $l(u)$ and $l(v)$ represent the labels of nodes $u$ and $v$, respectively.

These values serve to rank and select the nodes with highest heterophily rates that necessitate the addition of TE. Utilizing TE to assess the variance in features among chosen pairs of nodes, which typically belong to distinct classes, results in TE values that generally increase and stabilize after several epochs, correlating directly with the model's accuracy. This is because the convolution operator updates the features of the nodes based on their respective classes. 

Consequently, by correctly steering the variance in the features using formula (\ref{eq:teupdate}), TE amplifies the discriminative capability of the classification process. The output of the convolutional layers for a node $v_{i,j}$ is

\begin{equation}
\mathbf{H}_{i,j}=\mathbf{H}_{i,j} + max(TE_{Y_j\rightarrow X_i})
\label{eq:teupdate}
\end{equation}

Our method modifies the message aggregation mechanism of the convolutional operator in the GGCN algorithm. This involves a refined selection of nodes for alteration and identification of optimal values that boost the model's efficiency, while ensuring the sustainability of training duration. The changes and additions to the GGCN algorithm are outlined in \textbf{bold}. These changes modify the behaviour of the algorithm, according to a novel control mechanism. The main steps of the TE-GGCN training algorithm are:

\begin{enumerate}
   \item Apply the linear transformation on the input.
   \item Apply the Elu activation function.
   \item Successively apply for each convolutional layer:
   \begin{itemize}
     \item Apply the Elu activation function.
     \item Apply dropout.
     \item Compute and apply the scaling coefficient.
     \item Apply convolutional normalization.
     \item Compute the scaling coefficient for the sign matrices.
     \item Compute the sign attention matrices.
     \item Compute convolutional output with weights, sign matrices and coefficients.
   \end{itemize}
   \item \textbf{Compute the  heterophily rate using formula (\ref{eq:heterophily}}).
   \item \textbf{Select 5\% of the highest heterophilic nodes.}
   \item \textbf{From the above nodes subset select 10\% of highest degree nodes.}
   \item \textbf{Compute the TE values for the selected nodes.}
   \item \textbf{Compute the weights using formula (\ref{eq:teupdate}).}
   \item Apply softmax to obtain the classification outputs.
\end{enumerate}

Initially, we compute the hetherophily rate for all the nodes, after the convolutional layers are applied. We select 5\% of the initial nodes, the nodes with the highest hetherophily rate and also with highest degrees. The node selection count was determined  empirically after multiple runs, while trying  to optimally balance the computational cost and the accuracy.

Then we calculate the TE using equation (\ref{eq:TEcond}), constructing the $Y$ and $X$ series from the features of the selected nodes and their neighbors. After the convolutional layers have been applied using equation (\ref{eq:eq1}), we use the maximum calculated TE value for the selected nodes to increment the corresponding weights. Pre-selection of hetherophilic high-degreed nodes is required in order to keep the computational cost low and maximize the utility of our addition.

According to \cite{yan2022}, low-homophilic, low-degree nodes tend to change their class during training, but their influence among neighbors is directly correlated with their degrees, hence too small to influence the overall accuracy rate when TE mechanism is applied. Therefore, TE is not used for this particular group of nodes. The TE is calculated before the convolutional layers, since the weights and features do not influence the hetherophily rate before the convolutions.


\section{Experiments} \label{experiments}
\begin{figure*}[t]
\centering
\includegraphics[width=\textwidth,keepaspectratio]{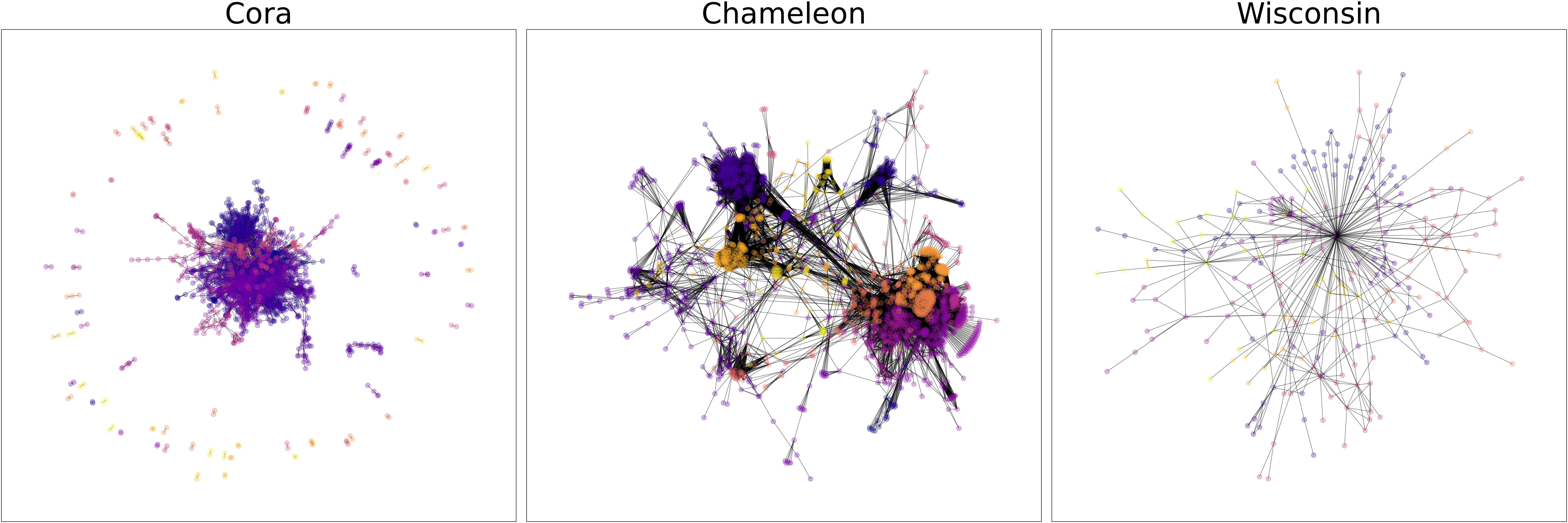}
\setlength{\abovecaptionskip}{-15pt}
\caption{Graph representation of selected datasets from different types of structure. Clustering was obtained using the Louvain method \cite{blondel2008} and colors do not represent the actual classes in the dataset. First, the Louvain algorithm converts the graph data into a k-nearest neighbor graph. Then, the degree (number of edges) from within a cluster is compared with the degree to the exterior of the cluster, hence computing weights for the edges between clusters. Cora corresponds to highly-homophilic graphs, Chameleon has both homophilic and heterophilic subgraphs, and Wisconsin presents heterophilic properties. It can be observed that Cora contains multiple disconnected subgraphs, whereas for Chameleon all subgraphs present low to high connectivity between the various classes.}
\label{fig:vargraphs}
\end{figure*}

To evaluate the performance of our TE-GGCN, we use real-world datasets, a well-known subset of the citation network type of datasets. These datasets express a variate range of both homophilic and heterophilic properties as well as combinations of these two. From the low homophilic datasets we use Chameleon and Squirrel \cite{rozemberczki2021}, Film (also known as Actor) from \cite{tang2009}, and from WebKB \cite{webkb1998} the Texas, Wisconsin, and Cornell datasets. From the high-homophilic datasets we use Cora \cite{mccallum2000}, CiteSeer \cite{sen2008}, PubMed \cite{yang2016}. A representative graphic of three of these datasets can be examined in Fig. \ref{fig:vargraphs}. 


In our study, we amplify the research presented in \cite{yan2022}, which encompasses a variety of GNN frameworks and demonstrates performance that is competitive with, or represents the state of the art for the selected datasets. Our objective is to address the challenge of finding the edge configurations that detrimentally impact classification accuracy in the later phases of training, commonly referred to as over-smoothing. To mitigate this issue, we employ TE to assess and subsequently modify the heterophily rate of a node, thereby enhancing its classification accuracy.

\subsection{Implementation and results}

Our TE-GGCN implementation uses PyTorch 2.1 and the Torch Geometric frameworks. It was derived from the \cite{yan2022} implementation and can be accessed and replicated through our Github\footnote{\url{https://github.com/avmoldovan/Heterophily_and_oversmoothing-forked/}} code.

The TE computation uses the implementation in \cite{pyif2020} since it provides the flexibility w.r.t. the inputs and also its sustained performance. This library relies on building K-D Trees as an intermediary step for the TE probabilities estimations. We evaluated that a window (or lag) of 1 is the best parameter value for obtaining TE values that can positively influence accuracy. Computing TE can generate extensive computational overheads given that some heterophilic nodes can have many neighbors, and it requires in the average scenario (low dimensionality and short inputs) a time in $O\left(n \log n+m \cdot n \cdot C_{\text {search }}\right)$, where $m$ is the number of nearest-neighbor searches per data point, $C_{\text {search }}$ is the cost of a single search, and $n$ is the number of datapoints. 

The TE-GGCN training time is influenced by the number of times the TE is computed. The number of TE calculations can vary considerably, depending on the degree of the selected nodes. For small degree nodes, the training cost is close to the original GGCN method, with less than 20\% execution time overhead. For nodes with high degree, even when selecting 5\% of their neighbors, TE computation adds a five fold overhead. We found that only the PubMed and Squirrel datasets require five, respectively four times more training time than the GGCN training method. For example, each of these these  datasets require more than 15M TE computations during a training session.

Table \ref{table:results} depicts the results of our TE-GGCN, compared with the ones reported in \cite{yan2022} for the GGCN model. For some datasets the GGCN results that we have obtained are slightly smaller than the ones reported by their authors in \cite{yan2022}, due to differences in our setup, which we extensively verified to match the ones of the original GGCN configuration from GitHub\footnote{\url{https://github.com/Yujun-Yan/Heterophily_and_oversmoothing}}. We note that when TE is computed and applied within \emph{each} of the convolutional layers, it offers a higher accuracy than the numbers reported for our TE-GGCN in Table \ref{table:results} on some datasets, but the computational overhead increases significantly (tens of times larger than the current one), making it impractical. Hence why we do not report these numbers here. 

\subsection{Discussion}
TE is a sensitive metric when used against the features of the nodes, since it can identify similar distributions and therefore similar connectivity patterns. Similar features generate smaller TE values. We can use TE values on the edges of a node disregarding its properties (homophilic or hetherophilic), in addition to the signed edge correction and decayed aggregation method \cite{yan2022}. Many node classification approaches attempted to use the homophily and hetherophily metrics in a systematic approach. We note that TE can independently influence the classification without the need to control these two metrics directly. 

Compared with GGCN, but also with more general GCN architectures, TE-GGCN offers an additional performance boost when classification accuracy is the main performance measure. In addition, increasing the number of nodes for which we compute TE also improves accuracy. As a trade off, this performance gain comes with a significant computational overhead when the TE is computed for all (or many) convolutional layers. Therefore, if higher accuracy is the main goal, the mechanisms that we implemented in TE-GGCN can be easily modified to squeeze extra performance from a GCN model.

\begin{table*}[htbp]
\centering
\caption{Datasets characteristics and mean accuracy over 10 runs with $\pm$ stdev. Best results are grayed and bolded.}
\resizebox{\linewidth}{!}{%
\begin{tabular}{|c|c|c|c|c|c|c|c|c|c|c}
\hline & Texas & Wisconsin & Actor & Squirrel & Chameleon & Cornell & Citeseer & Pubmed & Cora \\
\hline Hom. level $h$ & 0.11 & 0.21 & 0.22 & 0.22 & 0.23 & 0.3 & 0.74 & 0.8 & 0.81 \\
\hline Classes & 5 & 5 & 5 & 5 & 5 & 5 & 7 & 3 & 6 \\ 
\hline \#Nodes & 183 & 251 & 7,600 & 5,201 & 2,277 & 183 & 3,327 & 19,717 & 2,708 \\
\hline \#Edges & 295 & 466 & 26,752 & 198,493 & 31,421 & 280 & 4,676 & 44,327 & 5,278 \\
\hline TE-GGCN (ours) & \cellcolor{gray!15}\textbf{84.86 $\pm$ 4.55} & \cellcolor{gray!15}\textbf{87.45 $\pm$ 3.70} & $37.50 \pm 1.57$ & $55.04 \pm 1.64$ & \cellcolor{gray!15}\textbf{71.14 $\pm$ 1.84} & \cellcolor{gray!15}\textbf{85.68 $\pm$ 6.63} & \cellcolor{gray!15}\textbf{77.14 $\pm$ 1.45} & $89.08 \pm 0.37$ & \cellcolor{gray!15}\textbf{87.95 $\pm$ 1.05} \\
\hline GGCN & $83.51 \pm 3.72$ & $86.47 \pm 3.29$ & \cellcolor{gray!15}\textbf{37.56 $\pm$ 1.55} & \cellcolor{gray!15}\textbf{55.51 $\pm$ 2.06} & $70.57 \pm 1.84$ & $84.32 \pm 6.63$ & $76.51 \pm 1.45$ & \cellcolor{gray!15}\textbf{89.12 $\pm$ 0.32} & $84.32 \pm 1.05$\\
\hline
\end{tabular}
\label{table:results}
}
\end{table*}

\section*{Acknowledgment}
This article will be part of the 28th International Conference Information Visualisation (IV) - 2024 publications.

\section{Conclusions} \label{conclusions}
By leveraging node heterophily, degree metrics and TE values, we demonstrated that the GGCN method can be improved through straightforward methods. Utilizing TE as a measure of high node variances, we applied the highest TE value calculated in a forward pass as an adjustment to node features, post-convolution, without altering established mechanisms. This TE-based correction, applied prior to the softmax classification layer, aligns with our previous research \cite{Moldovan2021}, \cite{Moldovan2023}. Our approach is a versatile and easy way to improve existing GCN implementations, avoiding complex or tailored solutions.

\vspace{12pt}
\end{document}